\documentclass{article}
\usepackage{spconf,amsmath,graphicx}

\usepackage{amsmath}
\usepackage{bm}
\usepackage{tikz}
\usepackage{pgfplots}
\usepackage{xcolor}
\usepackage{bbm}
\usepackage{float}
\usepackage{adjustbox}

\definecolor{bblue}{HTML}{4F81BD}
\definecolor{rred}{HTML}{C0504D}
\definecolor{ggreen}{HTML}{9BBB59}
\definecolor{ppurple}{HTML}{9F4C7C}

\newcommand{\Real}{{\rm I\!R}}
\newcommand{\p}[1]{{\noindent \textbf{#1}}}

\renewenvironment{thebibliography}[1]{%
   \begin{oldthebibliography}{#1}%
     \setlength{\itemsep}{-.3ex}%
}%
{%
   \end{oldthebibliography}%
}

\title{Compressing local descriptor models for mobile applications}
\name{Roy Miles and Krystian Mikolajczyk}
\address{Imperial College London, UK}

\begin{document}

\maketitle

\begin{abstract}
Feature-based image matching has been significantly improved through the use of deep learning and new large datasets. However, there has been little work addressing the computational cost, model size, and matching accuracy tradeoffs for the state of the art models. In this paper, we consider these practical aspects and improve the state-of-the-art HardNet model through the use of depthwise separable layers and an efficient tensor decomposition.  
We propose the Convolution-Depthwise-Pointwise (CDP) layer, which partitions the weights into a low and full rank decomposition to exploit the naturally emergent structure in the convolutional weights. We can achieve an 8$\times$ reduction in the number of parameters on the HardNet model, 13$\times$ reduction in the computational complexity, while sacrificing less than $1\%$ on the overall accuracy across the \textit{HPatches} benchmarks. To further demonstrate the generalisation of this approach, we apply it to other state-of-the-art descriptor models, where we are able to a significant performance improvement.
\end{abstract}

\section{Introduction}
\label{sec:intro}

Local features have a wide range of applications in robotics, tracking, and 3D reconstruction, where the algorithms are often required to operate in real time on resource constrained devices. However, this is generally not possible for most CNN based models due to the memory and computational cost far exceeding the resources. Feature extraction and matching is a critical component in any virtual/augmented reality pipeline. 

Although the computational cost of handcrafted descriptors has been extensively researched \cite{Alcantarilla2013FastSpaces, Bay2008Speeded-UpSURF, Rosten2006MachineLF, Calonder2010BinaryFeatures}, there have been few methods for improving the efficiency of deep-learning based descriptors. In contrast, CNN models for image classification or object detection have been successfully compressed and deployed on mobile platforms through MobileNet \cite{Howard2017MobileNets:Applications} or ShuffleNet \cite{Zhang2018ShuffleNet:Devices}. This is not possible on the large models such as ResNet\cite{He2015ResNetRecognition}, VGG \cite{Simonyan2015VeryRecognition} or GoogLeNet \cite{Szegedy2015GoogLeNet/InceptionConvolutions}, which are commonly used as backbones for other tasks.

\begin{figure}
    \centering
    \adjustbox{width=0.8\linewidth}{
    \includegraphics{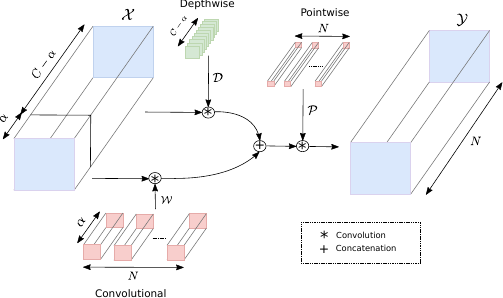}
    }
    \caption{The proposed Convolutional-Depthwise-Pointwise (CDP) layer partitions the input tensor across the depth. Most of the computational resources are then reserved for only a subset of the input features, while the rest use an efficient depth-wise convolution. The resulting features are then concatenated and aggregated using a pointwise convolution.}
    \label{fig:dw+conv}
\end{figure}
The number of parameters and the computational cost can act as a reasonable set of indirect metrics for the practical performance on-device, such as the inference latency. Our proposed method uses standard dense primitives, which have been efficiently implemented in most GPU-accelerated libraries, such as CuDNN.

We explore the use of popular low-rank decomposition methods \cite{Tucker1966SomeAnalysis,Hitchcock2015TheProducts} on two state-of-the-art descriptor models, namely HardNet \cite{Mishchuk2017WorkingLoss} and SuperPoint \cite{Detone2018SuperPoint:Description}. We provide an extensive evaluation of these descriptors with efficient operations and provide a practical scheme for combining them. 
Unfortunately, both depthwise-separable convolutions \cite{Sifre2014Rigid-MotionClassification} and Tucker decomposition \cite{Tucker1966SomeAnalysis} sacrifice the top-end accuracy of the models. 
To address this issue we propose a new layer, convolution-depthwise-pointwise (CDP), which partitions the input features to utilise both the standard convolution and the efficient depthwise convolution. The output features are then concatenated and aggregated using a pointwise convolution to maintain the original output dimensions (see figure \ref{fig:dw+conv}). Using this proposed decomposition, we can significantly compress the models with minimal degradation on the task accuracy.

\vspace{-0.8em}
\section{Related Work}
\label{sec:related_work}
The related work is divided into the recent advances of descriptor models, followed by the successful methods for compressing convolutional neural networks.

\subsection{Descriptors} 
\label{sec:related_work_descriptors}
There has been a lot of research on developing handcrafted descriptors that trade-off robustness for computational efficiency \cite{Rosten2006MachineLF,Bay2008Speeded-UpSURF,Calonder2010BinaryFeatures}.
However, machine learning approaches have been able to achieve significant accuracy improvements.
A number of recently proposed top performing descriptors have used the L2Net \cite{Tian2017L2-Net:Space} architecture with different training methodologies such as HardNet \cite{Mishchuk2017WorkingLoss}, GeoDesc \cite{Luo2018GeoDesc:Constraints}, SOSNet \cite{Tian2019Sosnet:Learning}, LF-Net~\cite{Ono2018LF-Net:Images}, etc. Despite the improved results of these descriptor models, they all leverage a large CNN backbone, which makes their deployment on mobile devices very difficult. To this end, we propose a drop-in replacement for the standard convolutional layer that proves to be very effective for models trained on image matching related tasks. To verify this claim, we consider the state-of-the-art HardNet \cite{Mishchuk2017WorkingLoss} and SuperPoint \cite{Detone2018SuperPoint:Description} models, whereby we are able to achieve significant model compression with minimal degradation in accuracy.

\vspace{-.8em}
\subsection{ CNN compression} 
\label{sec:cnn_efficiency_improvements}
This section presents several methods to improve the efficiency of the convolutional layers that have been successfully utilised in the context of object recognition, but not yet applied to local descriptors. 

\begin{figure*}[ht]
\begin{minipage}{.5\textwidth}
    \centering
    \includegraphics[width=0.75\linewidth,height=7em]{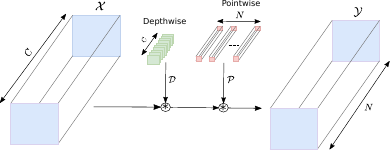}    
    \label{fig:dw_conv}
\end{minipage}%
\begin{minipage}{.5\textwidth}
    \centering
    \includegraphics[width=0.75\textwidth,height=7em]{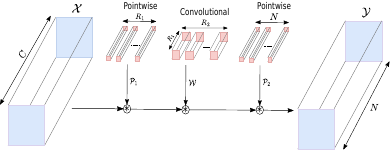}
    \label{fig:tucker}
\end{minipage}%
\caption{Depthwise separable convolution (left) and the Tucker decomposition applied to a convolutional layer (right)~\cite{Kim2015CompressionApplications}.}
\label{fig:dw_and_tucker}
\end{figure*}

%
%

\p{Pruning} is an active removal of individual weights, kernels, or even entire layers from a network based on a saliency measure or a regularization term. Optimal Brain Damage \cite{Lecun1990OptimalDamage} proposed to evaluate the saliency of individual weight entries using an approximation of the Hessian. This idea was further developed in Optimal Brain Surgery \cite{Hassibi1993SecondSurgeon} through iteratively computing the Hessian to obtain a more exact approximation.
\cite{Wen2016LearningNetworks, Li2017PruningConvnets} consider the pruning of weights in a group-wise fashion, while NISP~\cite{Yu2018NISP:Propagation} prune kernels through a propagated importance score, and \cite{Zhao2019VariationalPruning} use Bayesian inference with sparsity-inducing priors.
Pruning based methods for compression are orthogonal to low-rank decomposition, which is what we explore in our approach. 

\p{Low-rank tensor decomposition} is a method for approximating a higher order tensor using simple components, which are then combined using elementary operations. This can lead to a significant reduction in the number of parameters used to represent the original tensor, improve computational efficiency, and result in a more compact and interpretable model.
%
Depthwise-separable convolutions (see figure \ref{fig:dw_and_tucker} left) are the most common and have been used in all the MobileNet variants~\cite{Howard2017MobileNets:Applications, Fox2018MobileNetV2:Bottlenecks}. 
Xception~\cite{Chollet2017Xception:Convolutions} has also successfully used these layers as replacements in the Inception modules~\cite{Szegedy2015GoogLeNet/InceptionConvolutions}, however, they did not explore partitioning the input tensor across the depth, which is the focus of our proposed CDP layer. Tucker \cite{Tucker1966SomeAnalysis} and its special case Canonical Polyadic (CP) \cite{Hitchcock2015TheProducts} decomposition  factor an N-dimensional tensor into lower dimensional components.  The original convolution operation can then be replaced by a series of convolutions with smaller tensors  \cite{Jaderberg2014SpeedingExpansions, Kim2015CompressionApplications}. Tensor networks further provide a theoretical framework for such decomposition and have shown promising results \cite{Wang2018WideNets}.

\section{ Compressed Descriptor Model}
\label{sec:model_compression}
In this section we introduce our new CDP layer and a scheme that  combines the Tucker decomposition and depthwise-separable layers.
\subsection{Convolution-Depthwise-Pointwise (CDP)}
\begin{figure}[ht]
\centering
\adjustbox{width=0.65\linewidth,height=2.8em}{
\begin{tabular}{cccc}
      \includegraphics[width = 2.2in]{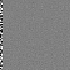} &
      \includegraphics[width = 2.2in]{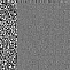} & 
\end{tabular}
}
\caption{Example of the convolutional weight slices $\mathcal{W}[:, :, i, j]$ from the pre-trained HardNet++ model for layers 6 and 7, Where $i, j$ is a given spatial coordinate in the receptive field. Note that the columns correspond to a given input channel index and the weights are scaled to be in the range $[0, 255]$.}
\label{fig:hardnetpp_weight_slices}
\end{figure}

Our proposed approach is motivated by the observation of two distinct partitions of the weights across the input channel depth. Figure \ref{fig:hardnetpp_weight_slices} shows some of the convolutional weight slices for the pre-trained HardNet++ model, where a significantly higher variance in the weight entries is present across only a subset of the input channels. In fact, we observe a clear and consistent cut-off point (offset) across each layer. These low-variance columns (input channel slices) can be approximated using a low-rank decomposition with a lower reconstruction error. In light of this, we enforce a partition on the convolutional weights that matches this observation before training.

Based on these observations, we propose an approach that combines a depthwise separable layer with a standard convolutional layer to provide a certain degree of full dense spatial and channel-wise connectivity for a subset of the input channels. This is shown in figure \ref{fig:dw+conv}, where the first few channels are reserved for a standard convolution, while depthwise kernels are used for the rest. These output features maps are then concatenated and aggregated using a pointwise convolution. We argue that, although the CDP layer will have more parameters than the typical depthwise separable layers, the dense connectivity will ensure that the high-end accuracy is maintained. We define the number of input channels for the standard convolution to be the offset parameter $\alpha$. This parameter provides a smooth transition between a normal convolution layer (albeit with a redundant pointwise convolution), where $\alpha$ is zero, and a fully depthwise separable layer, where $\alpha$ is equal to the number of input channels.

The input feature maps can be represented as a 3-way tensor, $\mathcal{X} \in \Real^{W \times H \times C}$, where $C$ indicates the number of feature maps and $W, H$ are the spatial dimensions. We use $\mathcal{X}_{i:j}$ to indicate feature maps indexed from $i$ (inclusive) through to $j$ (non-inclusive). This notation is used to define the operation of the CDP layer proposed. Let $\alpha \in [0, C]$ indicate the depthwise offset for compressing the convolutional layer and $\mathcal{W}, \mathcal{D}, \mathcal{P}$ correspond to the convolution, depthwise, and pointwise weights respectively.

\vspace{-1em}
\begin{align} 
    &\mathcal{Z}_{n} = \begin{cases} 
      \mathcal{X}_{0:\alpha} * \mathcal{W} & 0 \leq n < N \\
      \mathcal{X}_{\alpha:C} * \mathcal{D} & N \leq n < N + C
    \end{cases} \\[1.0ex]
    \medskip
    &\mathcal{Y} = \mathcal{Z} * \mathcal{P}
\end{align}

The total number of weights for stage (1) is given by \mbox{$K^2 \cdot \alpha \cdot N + K^2 \cdot (C - \alpha) $}, where $K^2$ is the receptive field size and $N$ is the number of kernels used for the standard convolution block. The output from both these blocks are then concatenated along the depth-axis and followed by a pointwise convolution with $O$ kernels. Both $\alpha$ and $N$ can be adjusted to control the overall compression however, for simplicity, we use $N = O$ throughout. On this basis, compression and acceleration of the overall layer is achieved if $\alpha < C - \frac{N}{K^2-1}$ (see Appendix).

\vspace{-.8em}
\subsection{Pointwise linear bottleneck}
\label{sec:pointwise_linear_bottleneck}
We also explore an approach that attempts to combine both Tucker decomposition and the depthwise separable layers. This method was motivated by the fact that the pointwise kernels contribute far more significantly to the total computation and the number of parameters than the depthwise kernels. 
We choose to replace the pointwise convolution in the depthwise separable layer with a bottleneck, where the size of this bottleneck is determined by Tucker decomposition with Variational Bayesian Matrix Factorization (VBMF)~\cite{Nakajima2013GlobalBayesian}. The core tensor is not used since the pointwise kernel has unit spatial dimensions and the maximum rank is chosen to ensure restorability of the models accuracy. In this case, model compression and acceleration is only achieved if the depth of the intermediate feature map $R$ is sufficiently small s.t. $CN > CR + RN$. This methodology differs from the Tucker decomposition  \cite{Kim2015CompressionApplications} by the fact that the core tensor is not used. 
This proposed scheme is inspired by the linear bottlenecks used in MobileNetV2~\cite{Fox2018MobileNetV2:Bottlenecks}, except that our spatial aggregation is not performed in this lower-dimensional subspace and the low-rank approximation is pre-computed using VBMF of the pre-trained network weights.

\section{Experimental results}
\label{sec:results}
The task performance of the descriptor models are evaluated using the HPatches benchmark \cite{Balntas2017HPatches:Descriptors}. This dataset is composed of local patches at varying levels of illumination and viewpoints changes from 116 different scenes. 
The HardNet model variants were all implemented in PyTorch \cite{Paszke2017AutomaticPyTorch} using the same training procedure\footnote{https://github.com/DagnyT/hardnet} and the HardNet++ weights were trained on the \textit{Liberty}, \textit{Yosemite}, and \textit{Notredame} datasets, while all the models proposed in this paper are trained solely on the \textit{Liberty} dataset from random initialisation.  

We perform an evaluation of the standard depthwise separable layers and Tucker decomposition on the HardNet model along with a comparison to our proposed pointwise linear bottleneck scheme and the use of CDP layers.

\vspace{-.8em}
\subsection{Performance metrics}

For our evaluation we consider both the computational cost, measured in the number of floating-point operations (FLOPs), and the total number of parameters across all the convolutional layers. Both the activation and batch normalisation operations are fused with the previous layers and so their cost has been omitted from calculations. 
The model size is described through a compression ratio i.e., the ratio of the total number of parameters in the original network against the compressed network. 

\vspace{-.8em}
\subsection{ Compressed HardNet performance}
We first report the baseline results for  state of the art descriptors and then compare the proposed accelerations  in terms of network compression ratio, mAP and computational cost (FLOPs). Table \ref{table:alternatives_hardnet} (top) compares the number of parameters and HPatches results for three descriptors frequently used in the literature. L2Net~\cite{Tian2017L2-Net:Space} and TFeat-M*~\cite{Balntas2017LearningNetworks} are CNN architectures and SIFT~\cite{Lowe2004SIFTKeypoints} is a handcrafted descriptor with square root normalisation~\cite{Arandjelovic2012ThreeRetrieval}. The SIFT consists of two convolutions to obtain image gradients which is equivalent to two 5x5 hardcoded kernels, thus 50 parameters

\begin{table*}[t]
    \centering
    \large
    \adjustbox{width=1.\linewidth}{
    \begin{tabular}{|c|c|c|c|c|c|c|c|c|c|c|c|c|c|c|c|c|c|}
        \hline
        \multicolumn{9}{|c|}{Layer offsets} &
        \multicolumn{3}{c|}{\;\;\;\;Homography Estimation\;\;\;\;} &
        \multicolumn{2}{c|}{\;\;\;\;Detector metrics\;\;\;\;} & 
        \multicolumn{2}{c|}{\;\;\;\;Descriptor metrics\;\;\;\;} &
        \multicolumn{2}{c|}{\;\;\;\;Performance\;\;\;\;} \\
        \hline
        \multicolumn{1}{|c}{\#2} & \multicolumn{1}{c}{\#3} & \multicolumn{1}{c}{\#4} & \multicolumn{1}{c}{\#5} & \multicolumn{1}{c}{\#6} & \multicolumn{1}{c}{\#7} & 
        \multicolumn{1}{c|}{\#8} & 
        \multicolumn{1}{|c}{\#9} & \multicolumn{1}{c|}{\#10} &
        \multicolumn{1}{c|}{$\epsilon = 1$} & 
        \multicolumn{1}{c|}{$\epsilon = 3$} & 
        \multicolumn{1}{c|}{$\epsilon = 5$} & 
        
        \multicolumn{1}{c|}{Rep.} & 
        \multicolumn{1}{c|}{MLE} & 
        \multicolumn{1}{c|}{NN mAP} &  
        \multicolumn{1}{c|}{M. Score} & 
        \multicolumn{1}{c|}{Compr.} & 
        \multicolumn{1}{c|}{Ops.} \\
        
        \hline
        \multicolumn{9}{|c|}{\textit{Original}} & 
        \multicolumn{1}{c|}{.440} &
        \multicolumn{1}{c|}{\textbf{.770}} &
        \multicolumn{1}{c|}{\textbf{.830}} &
        \multicolumn{1}{c|}{\textbf{.606}} &
        \multicolumn{1}{c|}{\textbf{1.14}} &
        \multicolumn{1}{c|}{.810} &
        \multicolumn{1}{c|}{\textbf{.550}} &
        \multicolumn{1}{c|}{$1\times$} &
        \multicolumn{1}{c|}{6.55} \\
        \hline
        
        \multicolumn{1}{|c}{2} & \multicolumn{1}{c}{4} & 
        \multicolumn{1}{c}{4} & \multicolumn{1}{c}{8} &
        \multicolumn{1}{c}{8} & \multicolumn{1}{c}{16} &
        \multicolumn{1}{c}{16} & \multicolumn{1}{|c}{16} &
        \multicolumn{1}{c|}{16} &
        \multicolumn{1}{c|}{.445} &
        \multicolumn{1}{c|}{.762} &
        \multicolumn{1}{c|}{.824} &
        \multicolumn{1}{c|}{.601} &
        \multicolumn{1}{c|}{1.06} &
        \multicolumn{1}{c|}{.835} &
        \multicolumn{1}{c|}{.519} &
        \multicolumn{1}{c|}{$2.58\times$} &
        \multicolumn{1}{c|}{2.10} \\
        \hline
        
        \multicolumn{1}{|c}{5} & \multicolumn{1}{c}{5} & 
        \multicolumn{1}{c}{5} & \multicolumn{1}{c}{5} &
        \multicolumn{1}{c}{5} & \multicolumn{1}{c}{5} &
        \multicolumn{1}{c}{5} & \multicolumn{1}{|c}{5} &
        \multicolumn{1}{c|}{5} &
        \multicolumn{1}{c|}{\textbf{.453}} &
        \multicolumn{1}{c|}{.752} &
        \multicolumn{1}{c|}{.822} &
        \multicolumn{1}{c|}{.596} &
        \multicolumn{1}{c|}{1.06} &
        \multicolumn{1}{c|}{.839} &
        \multicolumn{1}{c|}{.517} &
        \multicolumn{1}{c|}{$2.97\times$} &
        \multicolumn{1}{c|}{2.15} \\
        \hline
        
        \multicolumn{1}{|c}{2} & \multicolumn{1}{c}{2} & 
        \multicolumn{1}{c}{2} & \multicolumn{1}{c}{2} &
        \multicolumn{1}{c}{2} & \multicolumn{1}{c}{2} &
        \multicolumn{1}{c}{2} & \multicolumn{1}{|c}{2} &
        \multicolumn{1}{c|}{2} &
        \multicolumn{1}{c|}{.407} &
        \multicolumn{1}{c|}{.741} &
        \multicolumn{1}{c|}{.826} &
        \multicolumn{1}{c|}{.594} &
        \multicolumn{1}{c|}{1.06} &
        \multicolumn{1}{c|}{\textbf{.840}} &
        \multicolumn{1}{c|}{.511} &
        \multicolumn{1}{c|}{$\bm{3.21\times}$} &
        \multicolumn{1}{c|}{\textbf{1.91}} \\
        \hline
        
    \end{tabular}%
    }
    \caption{The CDP SuperPoint variants are able to achieve significant model compression with minimal degradation against the related detector and descriptor performance metrics. The computational complexity Ops. is measured in GFLOPs, and it is the contribution of the VGG16 backbone and the separate detector/descriptor heads.}
    \label{table:superpoint_cdp_layer}
\end{table*}

\p{Alternative methods}. To form a baseline for the evaluation of our proposed CDP layers, we consider the use of both depthwise separable layers, and an efficient Tucker decomposition implementation~\cite{Kim2015CompressionApplications}. We further consider the use of a pointwise linear bottleneck to combine the benefits of both these methods. The results can be seen in table~\ref{table:alternatives_hardnet}.

\begin{table}[h]
    \small
    \centering
    \adjustbox{width=1.\linewidth}{
    \begin{tabular}{|c|c|c|c|c|}
    
        \cline{1-4}
        Model & \# Parameters & Image Matching & Patch Retrieval & \multicolumn{1}{c}{} \\
        \cline{1-4}
        L2Net & 1,334,560 & \textbf{38.8} & \textbf{59.0} & \multicolumn{1}{c}{} \\ 
        \cline{1-4}
        SIFT & 50 & 25.7 & 42.7 & \multicolumn{1}{c}{} \\
        \cline{1-4} 
        TFeat-M* & 599,808 & 28.7 & 52.0 & \multicolumn{1}{c}{} \\
        \cline{1-4} 
        
        \multicolumn{5}{c}{} \\
    
        \hline
        Model & Compression ratio & Image Matching & Patch Retrieval & Operations (MFLOPs) \\
        \hline
        
        HardNet & $1 \times$ & \textbf{51.1} & \textbf{70.5} & 35.7 \\ 
        \hline
        DepthSep\{7\} & $4.3 \times$ & 50.1 & 69.4 & 34.6 \\ 
        \hline
        DepthSep\{6-7\} & $7.39 \times$ & 47.1 & 67.5 & 26.3 \\ 
        \hline
        DepthSep\{5-7\} & $10.72 \times$ & 46.4 & 67.0 & 25.4 \\ 
        \hline
        DepthSep\{2-7\} & $18.91 \times$ & 44.5 & 66.0 & \textbf{5.6} \\ 
        \hline
        \multicolumn{5}{c}{} \\
        \hline
        Tucker\{2-6\} & $1.21 \times$ & 50.7 & 70.1 & 10.3 \\  
        \hline
        Tucker\{7\} & $3.11 \times$ & 29.7 & 50.1 & 34.8 \\  
        \hline
        Tucker\{2-7\} & $6.81 \times$ & 21.8 & 42.2 & 9.4 \\  
        \hline
        
        \multicolumn{5}{c}{} \\
        \hline
        DepthSep\{7\} + Tucker\{2-6\} & $\bm{20.56 \times}$ & 27.4 & 47.3 & \textbf{8.5} \\ 
        \hline
        DepthSep\{7\} + TDW\{5-6\} & $12.01 \times$ & 47.0 & 67.2 & 25.0 \\  
        \hline
        
    \end{tabular}%
    }
    \caption{Comparison of using depthwise-separable layers, Tucker decomposition, and a proposed scheme for combining the two (TDW). The number(s) in the braces indicates the layers replaced. The image matching and patch retrieval accuracy is evaluated with mean Average Precision (mAP).}
    \label{table:alternatives_hardnet}
\end{table}

\begin{table}
    \centering
    \adjustbox{width=1.\linewidth}{%
    \begin{tabular}{|c|c|c|c|c|c|c|c|c|c|}
        \hline
        \multicolumn{6}{|c|}{Layer offsets} & \multicolumn{1}{c|}{Image Matching} & \multicolumn{1}{c|}{Patch Retrieval} & \multicolumn{1}{c|}{Compression ratio} & \multicolumn{1}{c|}{Operations}\\ 
        
        \hline
        
        \multicolumn{1}{|c}{\#2} & \multicolumn{1}{c}{\#3} & \multicolumn{1}{c}{\#4} & \multicolumn{1}{c}{\#5} & \multicolumn{1}{c}{\#6} & \multicolumn{1}{c|}{\#7} & \multicolumn{1}{c|}{mAP} & \multicolumn{1}{c|}{mAP} & \multicolumn{1}{c|}{} & \multicolumn{1}{c|}{MFLOPs}\\
        \hline
        \multicolumn{6}{|c|}{\textit{Original}} & \multicolumn{1}{c|}{\textbf{51.1}} & \multicolumn{1}{c|}{\textbf{70.5}} & \multicolumn{1}{c|}{$1\times$} & \multicolumn{1}{c|}{32.35}\\ 
        \hline
        
        %
        %
        
        \multicolumn{1}{|c}{2} & \multicolumn{1}{c}{2} & \multicolumn{1}{c}{2} & 
        \multicolumn{1}{c}{2} & 
        \multicolumn{1}{c}{2} & 
        \multicolumn{1}{c|}{2} & \multicolumn{1}{c|}{48.6} & \multicolumn{1}{c|}{68.6} & \multicolumn{1}{c|}{$\bm{9.50\times}$} & \multicolumn{1}{c|}{\textbf{10.65}} \\ 
        
        \multicolumn{1}{|c}{5} & \multicolumn{1}{c}{5} & \multicolumn{1}{c}{5} & 
        \multicolumn{1}{c}{5} & 
        \multicolumn{1}{c}{5} & 
        \multicolumn{1}{c|}{5} & \multicolumn{1}{c|}{50.3} & \multicolumn{1}{c|}{70.4} & \multicolumn{1}{c|}{$7.66\times$} & \multicolumn{1}{c|}{12.43} \\ 
        
        \multicolumn{1}{|c}{10} & \multicolumn{1}{c}{10} & \multicolumn{1}{c}{10} & 
        \multicolumn{1}{c}{10} & 
        \multicolumn{1}{c}{10} & 
        \multicolumn{1}{c|}{10} & \multicolumn{1}{c|}{50.0} & \multicolumn{1}{c|}{70.0} & \multicolumn{1}{c|}{$5.79\times$} & \multicolumn{1}{c|}{15.43} \\ 
        
        \multicolumn{1}{|c}{15} & \multicolumn{1}{c}{15} & \multicolumn{1}{c}{15} & 
        \multicolumn{1}{c}{15} & 
        \multicolumn{1}{c}{15} & 
        \multicolumn{1}{c|}{15} & \multicolumn{1}{c|}{50.4} & \multicolumn{1}{c|}{70.3} & \multicolumn{1}{c|}{$4.65\times$} & \multicolumn{1}{c|}{18.42} \\ 
        
        \hline
        
        %
        %
        
        \multicolumn{1}{|c}{2} & \multicolumn{1}{c}{4} & 
        \multicolumn{1}{c}{4} & 
        \multicolumn{1}{c}{8} & 
        \multicolumn{1}{c}{8} & 
        \multicolumn{1}{c|}{16} & \multicolumn{1}{c|}{50.0} & \multicolumn{1}{c|}{70.0} & \multicolumn{1}{c|}{$5.01\times$} & \multicolumn{1}{c|}{11.76} \\ 
        
        \multicolumn{1}{|c}{4} & \multicolumn{1}{c}{8} & 
        \multicolumn{1}{c}{8} & 
        \multicolumn{1}{c}{16} & 
        \multicolumn{1}{c}{16} & 
        \multicolumn{1}{c|}{32} & \multicolumn{1}{c|}{50.1} & \multicolumn{1}{c|}{70.1} & \multicolumn{1}{c|}{$3.21\times$} & \multicolumn{1}{c|}{14.07} \\ 
        
        \hline
        
        %
        %
        
        \multicolumn{1}{|c}{4} & \multicolumn{1}{c}{8} & 
        \multicolumn{1}{c}{8} & 
        \multicolumn{1}{c}{16} & 
        \multicolumn{1}{c}{16} & 
        \multicolumn{1}{c|}{2} & \multicolumn{1}{c|}{49.9} & \multicolumn{1}{c|}{69.9} & \multicolumn{1}{c|}{$7.61\times$} & \multicolumn{1}{c|}{13.83} \\ 
        
        \hline
    \end{tabular}%
    }
    \caption{Applying different depthwise offsets for the HardNet model with CDP layers. The results are evaluated on the HPatches benchmark and averaged over the Easy, Hard, and Tough distributions.}
    \label{table:cdp_layer}
\end{table}

\vspace{-.8em}
\p{CDP Layers.}
The results from table \ref{table:cdp_layer} demonstrate how enabling a subset of the channels to utilise full dense connectivity allows for the model to reach the high-end accuracy, while still achieving the favourable compression and acceleration from depthwise separable layers. The first group of rows considers a fixed offset for each layer, the second group takes into account the expansion of every odd layer by doubling the offset on each of these layers, and finally the bottom row uses the offsets as defined from the pre-trained HardNet++ weights (see figure \ref{fig:hardnetpp_weight_slices}). Specifically, for the last row, we define the offset as the channel index where the variance drops below the average across all channels. We observe that, as long as there is at least some  dense channel connectivity (i.e. $\alpha \geq 5$), the architecture is able to achieve the top-end accuracy, which was not attainable with just depthwise separable layers, Tucker decomposition, or even the proposed combined approach. The CDP variants are able to achieve the best balance between the number of parameters and computation, while also demonstrating very little drop in accuracy ($< 1\%$). Not only is the top-end accuracy maintained, but the model is pushed into a suitable range for real-time performance on low-compute mobile phones (10-150MFLOPs). 

\vspace{-.9em}
\subsection{SuperPoint}

We explore how the CDP layer can be applied to other models for a significant performance improvement. For this we consider the SuperPoint \cite{Detone2018SuperPoint:Description} model, which leverages a VGG \cite{Simonyan2015VeryRecognition} backbone to jointly predict the interest points and descriptors for matching. 

To ensure consistency, we follow the original training methodology for all the evaluations, which includes the same homographic adaptions of training images. The model is jointly trained using the pseudo ground truth labels on the MS-COCO~\cite{Lin2014MicrosoftContext} dataset, while the evaluation is performed using \textit{HPatches}~\cite{Balntas2017HPatches:Descriptors} benchmark.
The results for the complete image matching pipeline can be seen in table \ref{table:superpoint_cdp_layer} 
and show that the CDP layers are able to achieve a significant $\sim3.2\times$ reduction in parameters and a $\sim1.9\times$ reduction in FLOPs with minimal loss in the attainable matching score. 



\vspace{-.8em}
\section{Conclusions}
\label{sec:conclusion}
In this paper we demonstrate the accuracy/performance trade-offs of applying various factorisation and networks compression methods on CNN models used for local feature extraction.  We have proposed a novel Convolution-Depthwise-Pointwise (CDP) layer that consists of a partitioned low and full rank decomposition of the weights that matches the naturally emergent structure of the pre-trained weights. The allocated dense connectivity for a subset of the input features helps maintain the top-end descriptor accuracy. We further demonstrate the generalisability of this idea onto large architectures, namely the SuperPoint model. In both cases, we are able to compress the models significantly, with minimal to no accuracy degradation. This enables these models to be meet the resource constraints imposed by mobile devices for a wide host of applications, such as augmented/virtual reality. 



\bibliographystyle{IEEEbib-abbrev}
\bibliography{references_mendeley}
\end{document}